%% file: root.tex
\documentclass[letterpaper, 10 pt, conference]{ieeeconf} 

\IEEEoverridecommandlockouts                        % This command is only needed if 
\usepackage{graphics}
\usepackage{amsmath} 
\usepackage{amssymb}
\usepackage{cite}
\usepackage{tikz}
\usepackage{amsmath}
\usepackage{tabularx}
\usepackage{array}
\usepackage{booktabs}
\usepackage{algorithm}
\usepackage{algorithmic}
\usepackage{bbm}
\usepackage{adjustbox}
\usepackage{nicematrix}
\usepackage{caption}
\usepackage{gensymb}
\usepackage{svg}
\usepackage{flushend}
\usepackage{pifont}% http://ctan.org/pkg/pifont

\usepackage{tablefootnote}

\usepackage{hyperref}

\definecolor{green}{RGB}{11,155,13}

\newcommand{\dataset}[0]{M2P2}

\newcommand{\cmark}{\textcolor{green}{\ding{51}}} % ✓
\newcommand{\xmark}{\textcolor{red}{\ding{55}}}       
\definecolor{gmugreen}{RGB}{0, 102, 51}% ✗

\hypersetup{
    colorlinks=true,  
    urlcolor=gmugreen
    }



\title{\LARGE \bf
\dataset: A Multi-Modal Passive Perception Dataset for \\Off-Road Mobility in Extreme Low-Light Conditions
}

\author{Aniket Datar$^{*1}$, Anuj Pokhrel$^{*1}$, Mohammad Nazeri$^{*1}$, Madhan B. Rao$^{*1}$, Harsh Rangwala$^1$, Chenhui Pan$^1$, \\Yufan Zhang$^1$, Andr\'e Harrison$^2$, Maggie Wigness$^2$, Philip R. Osteen$^2$, Jinwei Ye$^1$, and Xuesu Xiao$^1$
\thanks{$^1$George Mason University {\tt\scriptsize \{adatar, apokhre, mnazerir, mbalajir, hrangwa, cpan7, yzhang82, jinweiye, xiao\}@gmu.edu}}
\thanks{$^2$DEVCOM Army Research Laboratory {\tt\scriptsize \{andre.v.harrison2.civ, maggie.b.wigness.civ, philip.r.osteen.civ\}@army.mil}}
\thanks{*Equally contributing authors}
}

\begin{document}
\maketitle
\pagestyle{empty}

%%%%%%%%%%%%%%%%%%%%%%%%%%%%%%%%%%%%%%%%%%%%%%%%%%%%%%%%%%%%%%%%%%%%%%%%%%%%%%%%
\input{content/abstract.tex}

%%%%%%%%%%%%%%%%%%%%%%%%%%%%%%%%%%%%%%%%%%%%%%%%%%%%%%%%%%%%%%%%%%%%%%%%%%%%%%%%
\input{content/intro.tex}

\input{content/related.tex}
\input{content/suite.tex}

\input{content/calibration.tex}
\input{content/dataset}
\input{content/experiment.tex}

\input{content/conclusion.tex}

% \newpage
\bibliographystyle{IEEEtran}
\bibliography{IEEEabrv,references}
\end{document}

%% file: content/abstract.tex
\begin{abstract}
Long-duration, off-road, autonomous missions require robots to continuously perceive their surroundings regardless of the ambient lighting conditions. 
Most existing autonomy systems heavily rely on active sensing, e.g., LiDAR, RADAR, and Time-of-Flight sensors, or use (stereo) visible light imaging sensors, e.g., color cameras, to perceive environment geometry and semantics. 
In scenarios where fully passive perception is required and lighting conditions are degraded to an extent that visible light cameras fail to perceive, most downstream mobility tasks such as obstacle avoidance become impossible. 
To address such a challenge, this paper presents a Multi-Modal Passive Perception dataset, \dataset, to enable off-road mobility in low-light to no-light conditions. 
We design a multi-modal sensor suite including thermal, event, and stereo RGB cameras, GPS, two Inertia Measurement Units (IMUs), as well as a high-resolution LiDAR for ground truth, with a multi-sensor calibration procedure that can efficiently transform multi-modal perceptual streams into a common coordinate system. 
Our 10-hour, 32 km dataset also includes mobility data such as robot odometry and actions and covers well-lit, low-light, and no-light conditions, along with paved, on-trail, and off-trail terrain. 
Our results demonstrate that off-road mobility and scene understanding under degraded visual environments is possible through only passive perception in extreme low-light conditions. 
The project website can be found at \url{https://cs.gmu.edu/~xiao/Research/M2P2/}.

\end{abstract}

%% file: content/intro.tex
\section{Introduction}
\label{sec::introduction}

Autonomous mobile robots have found their way out of controlled lab, factory, and warehouse environments into the wild~\cite{xiao2022motion}. On their way to deliver packages~\cite{hooks2020alphred}, inspect infrastructure~\cite{van2018mobile}, maintain agricultural fields~\cite{oliveira2021advances}, and conduct search and rescue missions~\cite{murphy2014disaster}, those robots constantly perceive their surroundings with their onboard sensors. The perceived geometric and semantic world representations allow them to move to their goals while avoiding collisions. Such an extension in Operational Design Domain requires robot perception systems to address challenges around the clock, ranging from well-lit to no-light conditions, as well as from paved to completely off-road terrain in the wild. 

Existing perception systems for mobile robots rely heavily on active sensing. 
For example, LiDAR range finders~\cite{wandinger2005introduction} use pulsed laser beams to detect distance and perceive environmental geometry, while Time-of-Flight sensors~\cite{li2014time} use infrared light and measure the time it takes for the light signal to travel to the target and back. 
Despite working well in all lighting conditions, many active sensors suffer from significant noise in heavy rain, snow, and fog. Furthermore, the reliance on the emission of active light signals will expose the presence of the robot, making those active sensors less ideal for covert operations, e.g., in military settings. 

\begin{figure}
  \centering
  \includegraphics[width=\columnwidth]{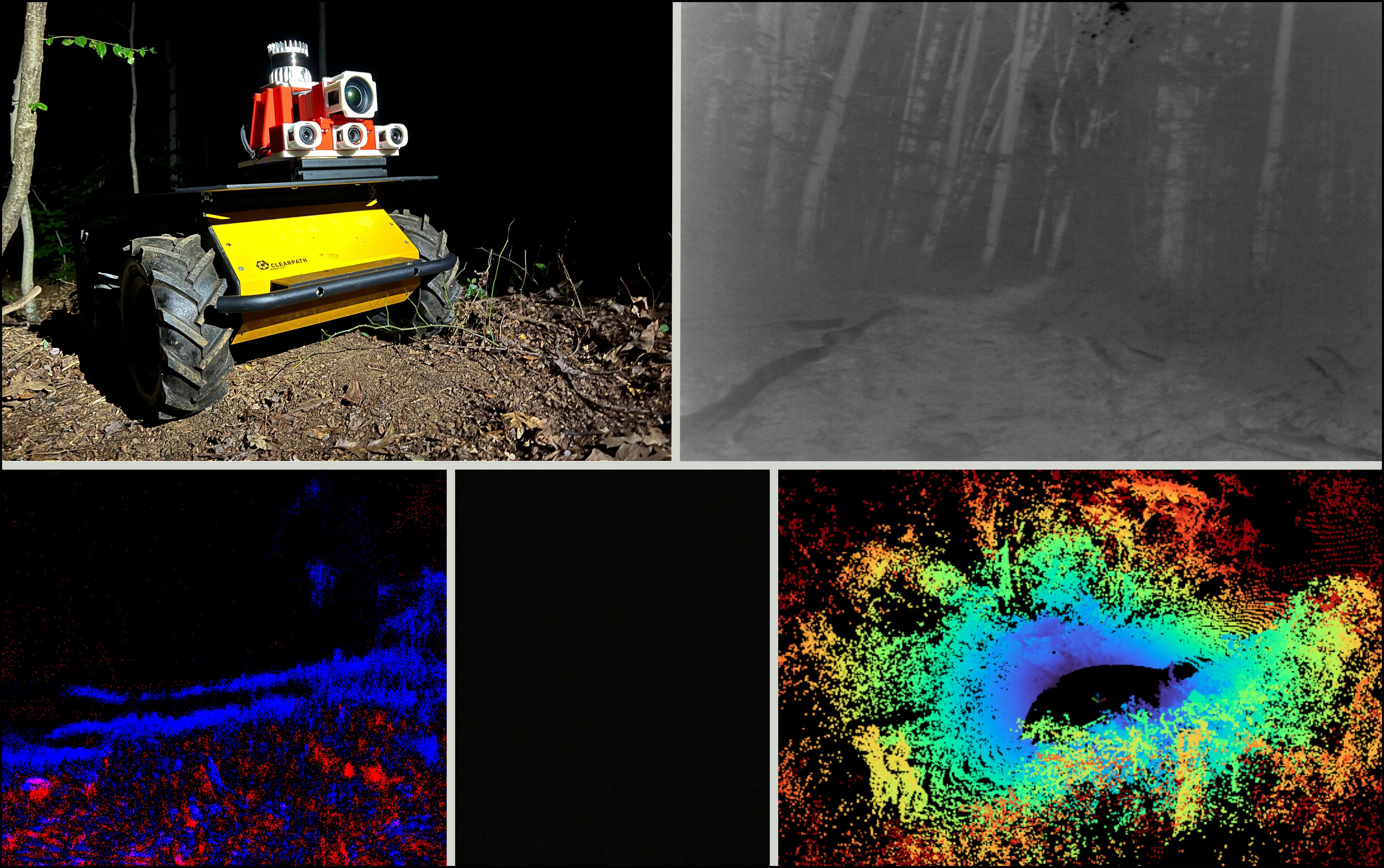}
  \caption{Multi-Modal Passive Perception Data Collection in an Off-Road Forest Environment in Complete Darkness. Top Left: Clearpath Husky with the Sensor Suite (flashlight for visualization only); Top Right: Thermal Image; Bottom Left: Event Stream; Bottom Middle: RGB Image (fail to perceive); Bottom Right: LiDAR Point Cloud (for ground truth).}
  \label{fig::overview}
\end{figure}

Non-active, visible light imaging sensors, e.g., RGB cameras, are also widely used in robot perception systems,  relying on reflected light to form images for non-light emitting objects. Stereo camera pairs can triangulate to determine distance and use different RGB color channels to reason about semantics. Those sensors work well in well-lit indoor and outdoor environments and provide similar sensing as human perception. However, visible light
imaging sensors require good lighting conditions to perceive reflected light and form visible pixels, and therefore suffer from degraded perception quality in low-light to no-light conditions. 

These aforementioned limitations of existing active and visible light imaging sensors present challenges for long-duration, off-road, autonomous missions, since robots need to perceive their surroundings around the clock regardless of the ambient lighting conditions and are also oftentimes required to be fully passive to maintain stealth. To operate in low-light to no-light conditions without emitting any active light signatures, novel sensing modalities, including thermal and event cameras, show promise by passively sensing infrared radiation from all objects with a temperature above absolute zero or per-pixel brightness changes (also called ``events'') asynchronously with low latency, high dynamic range, and low power consumption, respectively. 

In this paper, we propose to use multi-modal passive perception modalities to enable robot perception in extreme low-light conditions so as to facilitate downstream off-road mobility tasks (Fig.~\ref{fig::overview}). To be specific, our contributions include: 
\begin{itemize}
    \item a multi-modal sensor suite including thermal, event, and stereo RGB cameras, GPS, two IMUs, and a high-resolution LiDAR for ground truth;
    \item a precise multi-sensor calibration procedure for multi-modal perceptual streams;
    
    \item a Multi-Modal Passive Perception dataset, \dataset, with data ranging from different lighting conditions (well-lit to no-light) and various off-road terrain conditions (paved to off-trail), along with mobility data like robot odometry and actions; and 
    \item experimental results demonstrating off-road mobility, depth reconstruction, and vehicle odometry through only passive perception in extreme low-light conditions. 
\end{itemize}

%% file: content/related.tex
\section{Related Work}
\label{sec::related_work}

In this section, we review related work in off-road perception systems and passive perception sensors. 

\subsection{Off-Road Perception}
Perception in off-road environments requires both exteroceptive and interoceptive sensing to understand the environment and the robot's interaction with it.
The availability of a wide array of sensors makes safe traversal through off-road environments possible.
While a single modality may suffice for navigation in structured environments, the inclusion of multiple modalities in challenging environments adds robustness and redundancy, 

ensuring that navigation can continue even if one or more sensors are unable to work at full capacity because of adverse environmental conditions.
By combining complementary data from multiple sensors, robots can also better perceive and interpret complex environmental features for comprehensive understanding in a variety of off-road unstructured scenarios. 

%%%%%%%%%%%%%%%%% adding dataset comparison table
\input{content/comparison_table}

Active sensing modalities like LiDAR and RADAR detect and perceive environmental geometry, enabling the creation of 2D, 3D, or 2.5D elevation maps~\cite{ebadi2020lamp, thakker2021autonomous, lamp2, fankhauser2016navigation, RAMP2023} of the environment.
Although LiDAR-based systems are highly popular for their robustness and precision, they can suffer in heavy rain, snow, and fog, and may struggle to map terrain at greater distances~\cite{chung2024pixel}. Additionally, the use of pulsated beams can expose the presence of the robot.
On the other hand, vision-based navigation systems utilize visible light imaging sensors, e.g., RGB or RGB-D cameras, to understand the terrain semantics~\cite{whereshouldiwalk, fankhauser2018terrain, rugd2019dataset, meng2023terrainnet}, create elevation maps~\cite{chung2024pixel, fankhauser2018terrain}, and map off-road terrain~\cite{sermanet2008, bajracharya2013high}.
Although vision-based navigation systems are advantageous due to their passive sensing capabilities and ability to provide rich environmental information, their reliance on visible light causes poor performance in low-light conditions. 

While also being passive, interoceptive sensors like IMUs and force sensors measure robot internal states during environment interactions, which can be used to generate traversability maps~\cite{whereshouldiwalk, howdoesitfeel} and model terrain response~\cite{cai2022risk, kahn2021badgr} when combined with exteroception.

Combining the advantages of the aforementioned perception modalities expands robots' Operational Design Domain in varying environmental conditions around the clock, such as low visibility or extreme weather, with the possibility of staying passive.
With the recent advancement in data-driven approaches~\cite{xiao2022motion},  multi-modal off-road datasets~\cite{diter, jiang2021rellis, lee2022vivid++} are essential for developing and refining perception and mobility algorithms, providing a foundation for training, testing, and benchmarking.
Our multi-modal sensor suite offers passive sensing capabilities with precise ground truth from active perception, enabling navigation in extremely low-light off-road environments.
The sensor suite is resilient to environmental degradation like dust, smoke, fog, snow, and rain, and can be calibrated in a single step for effective off-road navigation.

\subsection{Related Datasets}
A few existing datasets provide a variety of sensor modalities and ground truth data, enabling the development and benchmarking of algorithms in areas such as SLAM, object recognition, and autonomous navigation (Table \ref{tab:comparison}):
MVSEC~\cite{MVSEC} is the first dataset that synchronizes stereo event cameras and provides accurate ground truth depth from LiDAR and SLAM and ground truth pose using a motion capture system and GPS; UZH-FPV~\cite{UZH-FPV} dataset utilized fast, aggressive, and agile drones to capture event camera data for extreme motion scenarios, but does not contain depth information; For night and day place recognition tasks, Maddern and Vidas~\cite{Maddern2012TowardsRN} built a capture platform consisting of GPS, RGB camera, and thermal camera to capture data from before dawn to after dusk; The KAIST Multi-Spectral Day/Night Dataset~\cite{KAIST} introduced a sensor system designed for SLAM, comprising stereo RGB cameras, LiDAR, and thermal camera; Aiming at off-road environments such as forests and urban areas, M3ED~\cite{M3ED} used high resolution stereo event cameras, grayscale and RGB cameras, IMU, LiDAR, and RTK localization to collect a high-speed dynamic motion dataset; ViViD++~\cite{lee2022vivid++} is the first dataset to feature aligned information from multiple types of alternative vision sensors, including RGB, thermal, event, depth, and inertial measurements. Compared to existing datasets, our M2P2 dataset is the first dataset that focuses on off-road mobility in extremely low-light environments with the most perception modalities and highest sensor quality, as well as a precise multi-modal calibration procedure with accurate synchronization (see Table \ref{tab:comparison} for comparison).  

%% file: content/comparison_table.tex
\begin{table*}[t]
\centering
\caption{Comparison with alternative vision datasets.}\label{tab:comparison}

\begin{tabular}{lccccccccccc}

\toprule                                                
   & \multicolumn{7}{c}{\textbf{Sensor Modality}} &  &  &\\
\cmidrule(rl){2-8}
\textbf{Dataset}                      & \textbf{RGB} & \textbf{Depth} & \textbf{Thermal} & \textbf{Event} & \textbf{LiDAR} & \textbf{IMU} & \textbf{GPS} & \textbf{Hardware} & \textbf{Environments} & \textbf{Lighting} \\
\midrule
ViViD++~\cite{lee2022vivid++}          & \cmark   & \cmark     & \cmark   &  \cmark  &  \cmark  & \cmark  & \cmark & Vehicle & Indoor/Urban & Day/Night\\ 
DiTer++~\cite{kim2024diter++}          & \cmark   & \cmark     & \cmark   &  \xmark  &  \cmark  & \cmark  & \cmark & Legged & Diverse Terrain & Day/Night\\
TartanDrive 2.0~\cite{tartandrive}     & \cmark   & \cmark     & \xmark  &    \xmark & \cmark  & \cmark & \cmark & Wheeled & Off-road & Day\\
\midrule
M2P2          & \cmark   & \cmark     & \cmark       &  \cmark    &  \cmark & \cmark & \cmark & Wheeled & Off-road & Day/Night\\

\bottomrule
\end{tabular}
\end{table*}

%% file: content/suite.tex
\section{Multi-Modal Sensor Suite}
\label{sec::suite}
\begin{figure}
    \centering
    \includegraphics[width=1\linewidth]{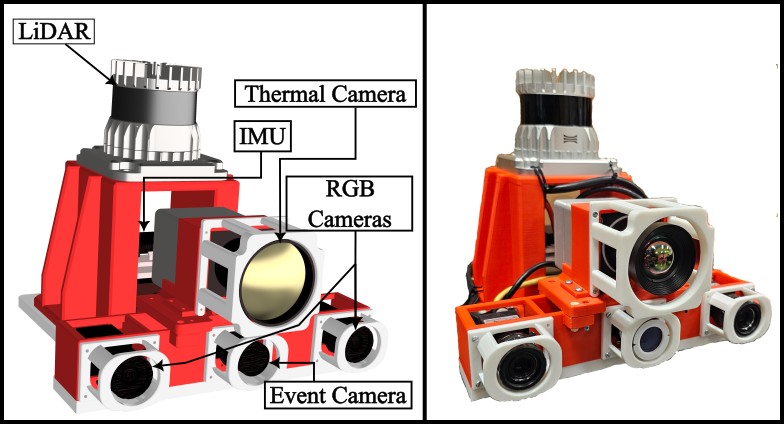}
    \caption{Sensor Suite CAD (Left) and Hardware (Right).}
    \label{fig::cad_comparision}
    \vspace{-15pt}
\end{figure}

Our multi-modal sensor suite comprises a thermal and an event camera, stereo RGB cameras, two IMUs, GPS, and LiDAR for ground truth. All sensors are assembled in a custom-designed 3D-printed structure, which can be easily mounted on most mobile robot platforms (Fig. \ref{fig::cad_comparision}). The total dimensions of the sensor suite are 0.31$\times$0.26$\times$0.24 m, with a total weight of 2 kg. 

\subsection{Thermal Camera}
Our sensor suite includes a Xenics Ceres T 1280 thermal camera, which features Long Wave Infrared (LWIR) imaging at a high resolution of 1280$\times$1024.
The camera can capture images at a maximum of 45 FPS via the GigE Vision interface.
The thermal camera is paired with a wide-angle lens of 11 mm with 71.7$\degree$ Horizontal Field of View (HFoV), 58.9$\degree$ Vertical FoV (VFoV), and an aperture of f/1.2. Notice that our wide-angle LWIR camera provides the highest quality thermal images compared to any existing open-source datasets. 

\subsection{Event Camera}
We use a Prophesee Metavision EVK4 as our event camera. 
The camera has a latency of 220 $\mu$s within a compact size with a sensor resolution of 1280$\times$720. 
We use a lens with 46.8$\degree$ HFoV and 36$\degree$ VFoV with an aperture range from f/2-11 (fixed at f/4.0).
The camera has a time resolution equivalent to 10K FPS and a low-light cutoff of 0.08 lx.
To prevent LiDAR pulses from introducing noisy events, we apply an IR filter in front of the event camera lens. 

\subsection{Stereo RGB Cameras}
We use two FLIR Blackfly S cameras for capturing images in the RGB spectrum.
The cameras have a resolution of 1616$\times$1240, which can be captured at a maximum of 175 FPS (fixed at 10 FPS). 

While our stereo RGB cameras fail to perceive in no-light conditions, they can still perceive in environments featuring only partial degradation or with some ambient lighting. 

\subsection{IMUs}
We use a Yahboom 10-DoF IMU featuring a 3-axis accelerometer, 3-axis gyroscope, 3-axis magnetometer, and a barometer. 
The sample rate of the IMU is 200 Hz.
It features built-in data fusion and gyro stabilization.

We also include the IMU embedded in the LiDAR (see details below). 

\subsection{LiDAR for Ground Truth}
A 3D Ouster OS1-128 LiDAR is used to provide ground truth with 128 lines of vertical divisions in 45$\degree$ VFoV and selectable 512, 1024, and 2048 angle divisions in 360$\degree$ HFoV at 10/20 Hz. 
For best data efficiency, LiDAR point clouds are recorded with 1024 angle divisions at 10 Hz. 
The LiDAR also features a built-in 6-DoF IMU with a 125 Hz sample rate for LiDAR frame calibration.

%% file: content/calibration.tex
\section{Sensor Suite Calibration}
\label{sec::calibration}

To understand how the multi-modal perception streams from the sensor suite transform real-world features in world coordinates into their corresponding sensor readings, as well as how they correlate with each other in terms of a common coordinate system, we develop a streamlined multi-modal calibration procedure to calibrate all the sensors with different modalities in the sensor suite. 

Traditional calibration methods use distances measured by geometric features, such as a printed black and white checkerboard with squares of known sizes for camera intrinsics and camera-to-camera extrinsics calibration, or a flat surface for LiDAR-to-camera extrinsics calibration. 
However, for our multi-modal sensor suite, those methods pose a limitation as conventional calibration targets are not visible in the infrared range of a thermal camera.
Furthermore, static calibration targets are not visible by an event camera, which needs motion to detect the changes in intensity. 

Therefore, our multi-modal sensor suite requires a common calibration target that can be perceived by all sensors as to calibrate both intrinsic and extrinsic parameters.

\subsection{Thermal Checkerboard}
\begin{figure}
    \centering
    \includegraphics[width=1\linewidth]{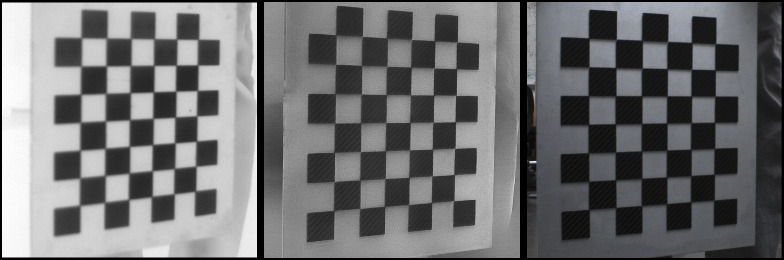}
    \caption{Calibration Target (Thermal, Event, and RGB Image).}
    \label{fig::target}
    \vspace{-10pt}
\end{figure}

The first challenge of calibrating our sensor suite comes from the thermal camera, which requires different thermal signatures to reflect distances of geometric features. 
To introduce a contrast thermal signature, we create a calibration target using an aluminum sheet of 3 mm thickness and carbon fiber squares of 35 mm. The sheet and the carbon fiber squares are precision milled with CNC achieving an accuracy of 0.05 mm.
Since the aluminum sheet is highly reflective in the long wave infrared (IR) spectrum (similar to a mirror in the visible spectrum), 
we anodize the aluminum sheet to eliminate unwanted reflection in the IR spectrum. After heating the calibration target to roughly 45$\degree$C, due to a large difference in emissivity of aluminum and carbon fiber, the checkerboard pattern appears in the thermal image (Fig.~\ref{fig::target} left). 
Due to the contrast in color of aluminum and carbon fiber, the same pattern is visible in both RGB cameras (Fig.~\ref{fig::target} right).

\subsection{Event Reconstruction}

To address the second calibration challenge of correlating asynchronous event data with other synchronous data streams, such as thermal and RGB images, we employ a two-step approach. First, we reconstruct a grayscale image from the raw event stream using E2Calib~\cite{Muglikar2021CVPR} (Fig.~\ref{fig::target} middle). Additionally, we utilize the trigger input functionality of the event camera to precisely mark timestamps for frame reconstruction, enabling accurate temporal alignment between the reconstructed event frames and corresponding frames from other sensors. This method allows us to overcome the inherent asynchronous nature of event data and establish reliable temporal relationship with synchronous data streams, facilitating multi-modal sensor fusion and calibration.

\subsection{Multi-Modal Synchronization}
\begin{figure}
    \centering
    \includegraphics[width=0.8\linewidth]{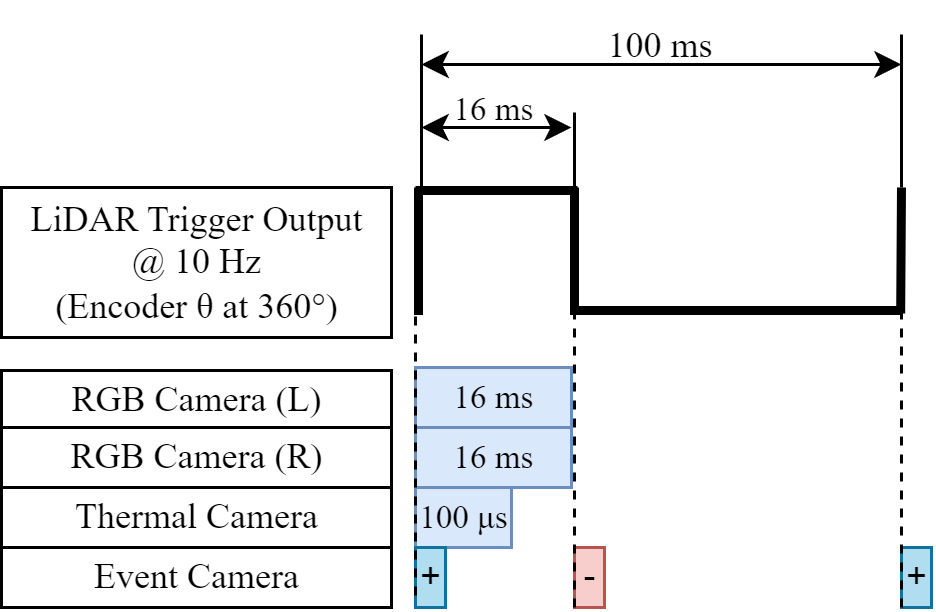}
    \caption{Multi-Modal Synchronization: LiDAR trigger synchronized to internal encoder angle ($\theta=360\degree$) initiates frame acquisition at a rate of 10 Hz for RGB and thermal cameras, with event camera recording trigger edges for frame reconstruction.}
    \label{fig::trigger}
    \vspace{-10pt}
\end{figure}

With a common calibration target visible in all four cameras in the sensor suite, with another RGB camera in the stereo pair, the last calibration challenge is the precise synchronization among multiple asynchronous and un-synchronized data streams to achieve calibration convergence. 
To address this, we implement a synchronization scheme as illustrated in Fig.~\ref{fig::trigger}.

We synchronize all four cameras to the LiDAR, which generates a 10 Hz sync pulse aligned to its encoder angle at 360\textdegree. This pulse triggers frame acquisition in the RGB and thermal cameras, with its edges marking temporal points in the event camera stream. The pulse width matches the RGB camera's exposure time, and its falling edge is used for event camera frame reconstruction. This approach aligns the reconstructed frame with the RGB camera's exposure completion, ensuring precise temporal correlation across all sensors.

\subsection{All-in-One Calibration Procedure}
\begin{figure}
    \centering
    \includegraphics[width=0.8\linewidth]{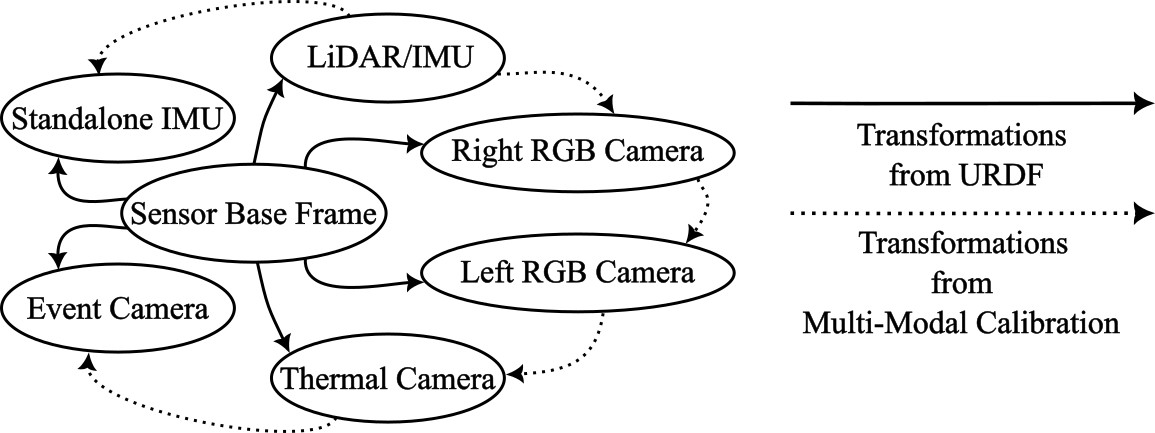}
    \caption{Transformation Tree of the Sensor Suite:  Solid arrows indicate direct hardware transformations, while dotted arrows represent transformations from our multi-modal calibration.}
    \label{fig::tree}
    \vspace{-15pt}
\end{figure}

Finally, we splice all the synchronized frames and create a ROS-bag that can be used with any calibration toolkit. In our implementation, we use Kalibr~\cite{furgale2013unified} calibration toolkit to generate camera intrinsic and extrinsic parameters. Furthermore, we need to calibrate the camera and IMUs to complete the transformation tree for the entire sensor suite. As the Ouster IMU features a 6-DoF IMU with factory-calibrated transformation from the LiDAR base to the IMU frame, we use the Ouster base as a reference frame to bind everything into a single tree. 
The entire transformation tree of the sensor suite from our multi-modal calibration, as well as from our hardware design, is shown in Fig.~\ref{fig::tree}.

\begin{figure}
    \centering
    \includegraphics[width=1.0\linewidth]{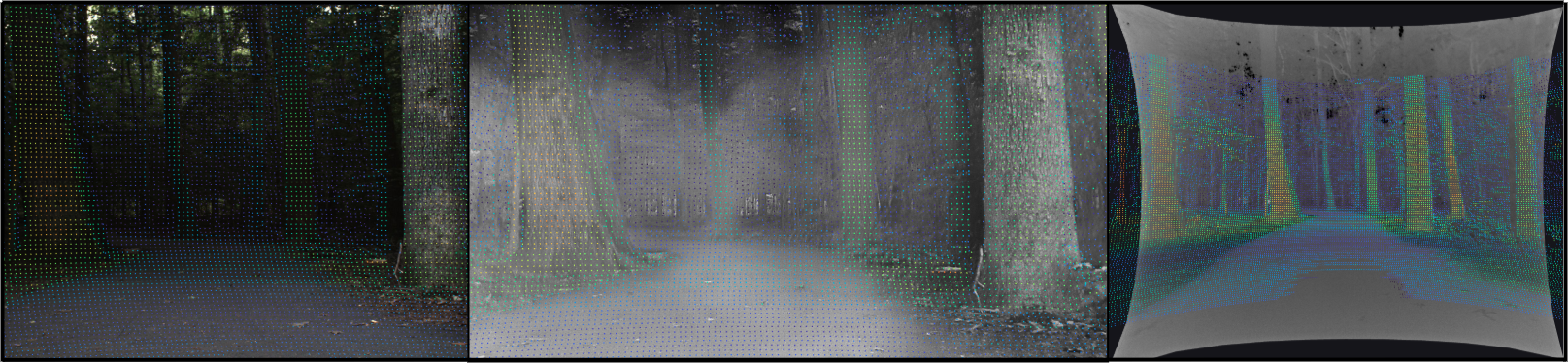}
    \caption{Multi-modal data from the M2P2 dataset, showcasing spatial and temporal alignment in a low-light, off-road forest environment. LiDAR point cloud overlaid on RGB image (left), reconstructed event frame at the trigger's falling edge (middle), and thermal image (right).}
    \label{fig::lidar_to_img}
    \vspace{-10pt}
\end{figure}

Fig.~\ref{fig::lidar_to_img} shows the LiDAR point cloud overlaid on the corresponding RGB image, along with the reconstructed event frame and thermal image, demonstrating the spatial and temporal alignment of the multi-modal data.

%% file: content/dataset.tex
\section{Multi-Modal Passive Perception Dataset}

\begin{figure*}
  \centering
  \includegraphics[width=2\columnwidth]{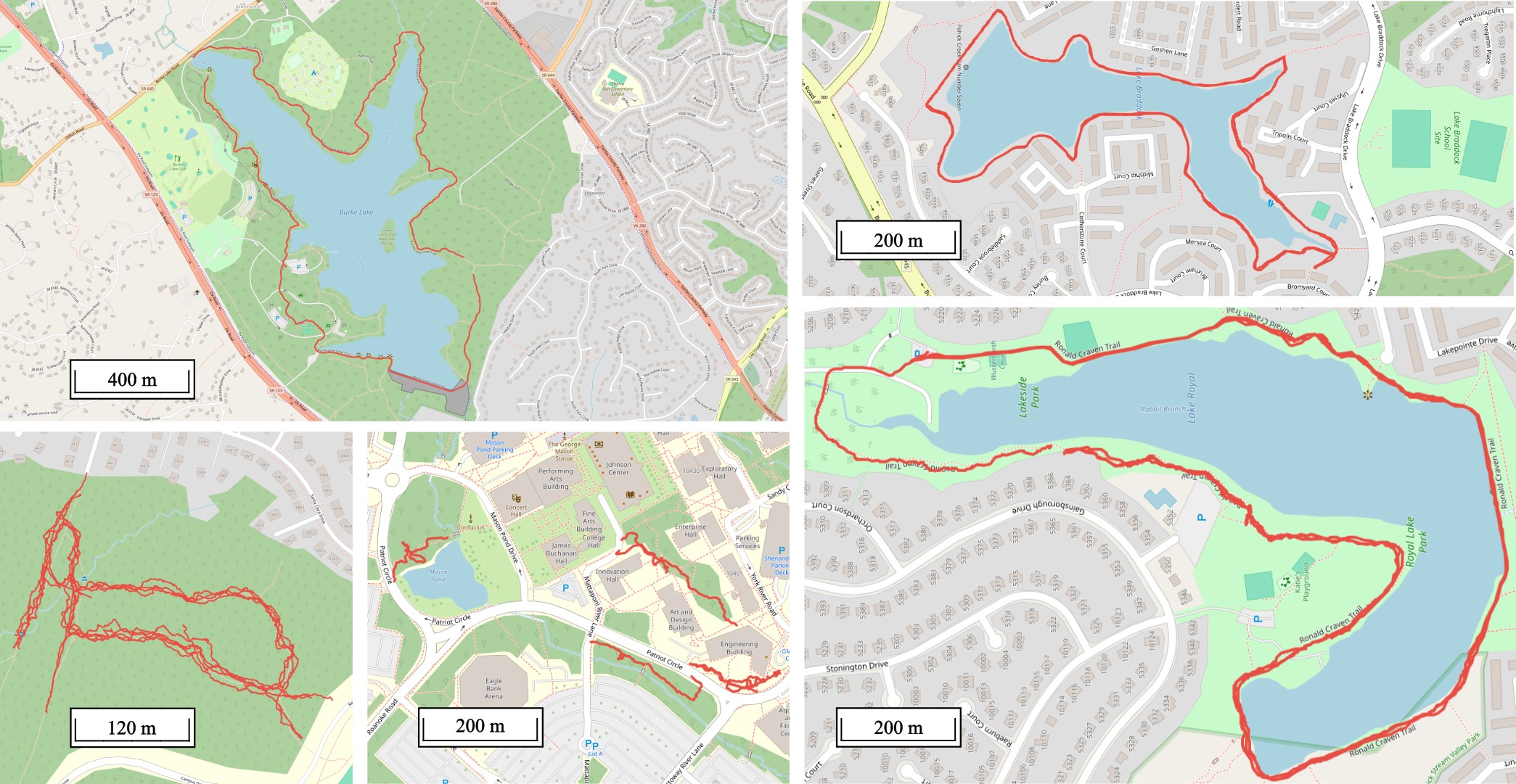}
  \caption{$>$32 km, 10.15-Hour M2P2 Data Collection across Different Locations: Maps show diverse environments including lakeside trails, urban parks, and dense forests, highlighting the variety of terrain and conditions captured in the dataset.}
  \label{fig::data_collection_locations}
  \vspace{-10pt}
\end{figure*}

\dataset~dataset encompasses over 10 hours of data collected across various challenging terrain conditions (Fig.~\ref{fig::data_collection_locations}). The data are gathered with the sensor suite mounted on a Clearpath Husky A200 robot. The dataset includes sequences from a diverse range of environments, progressing from fully prepared paved trails to non-paved off-road paths, and ultimately to unprepared off-trail environments within densely forested areas featuring thick vegetation and narrow passages. To capture a comprehensive range of lighting conditions, data collection is conducted at dusk, with luminosity levels varying from 20 lx to complete darkness (0 lx). This approach ensures the dataset's applicability to both well-lit and no-light scenarios, addressing the challenges of navigation in varying environmental conditions.

\begin{figure}
    \centering
    \includegraphics[width=0.9\linewidth]{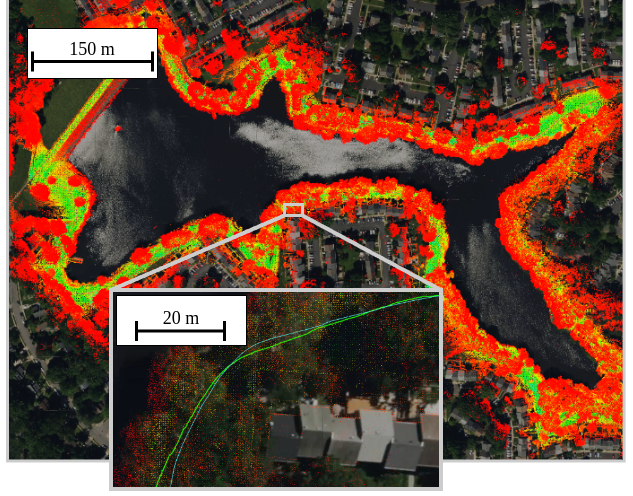} % Placeholder image
    \caption{LIO-SAM Mapping Results on Lake Braddock Trail: The LiDAR point cloud (colored points) is overlaid on a satellite image. Inset: Comparison of LIO-SAM estimated trajectory (blue) and raw GPS trajectory (green).}
    \label{fig::lake_braddock_map}
    \vspace{-10pt}
\end{figure}

The dataset is structured as ROS-bag files, consisting of compressed RGB and thermal images at 10 FPS, asynchronous raw event stream, 3D point cloud data from LiDAR, IMU data, GPS coordinates, robot odometry and status messages, and human-commanded joystick inputs. All camera data are synchronized using the trigger pulse from the LiDAR, ensuring temporal alignment across multi-modal sensor inputs. Due to the dense canopy of the trees the GPS data is only available for 87.97\% of the total dataset. However, it is possible to fuse LiDAR, IMU, and GPS, when available, with LIO-SAM~\cite{liosam2020shan}, relying primarily on lidar-inertial odometry. Fig.~\ref{fig::lake_braddock_map} shows a LIO-SAM-generated map overlaid on a satellite image. The LiDAR point cloud aligns well with visible features (e.g., trail edges and vegetation), demonstrating mapping accuracy. The inset compares the estimated trajectory (blue) to raw GPS (green); the latter deviates significantly under dense tree cover, reflecting degraded signal quality, while the LIO-SAM trajectory remains consistently accurate. 

To facilitate accurate sensor placement replication, we provide the URDFs (Unified Robotics Description Format) for the sensor suite configuration on the Husky platform, along with the calibrated transformations.
Table \ref{tab::dataset} shows the main statistics of the M2P2 dataset. 
The multi-modal synchronization scheme achieves near-perfect alignment between RGB images and LiDAR point clouds, with only six instances of mis-synchronization. The slightly reduced number of thermal images compared to RGB images is due to the thermal camera's automatic shutter calibration, which interrupts the image stream for approximately 0.4 seconds to correct for non-uniformities. 

\begin{table}
\centering
\caption{\dataset~Statistics}
\begin{tabular}{cc}
\toprule
 Attribute & Quantity \\
 \midrule
 Total Size & $\approx$2 TB \\
 Total Distance & $>$32 km\\
 Total Time & 10.15 h\\
 Total GPS Lock Time & 8.93 h\\
 % Average GPS Accuracy & \textcolor{red}{3.58 m}\\
 Average Speed & 0.95 m/s \\
 Number of RGB Images & 730606\\
 Number of Thermal Images & 361685\\
 Number of Events & $1.15\times10^{11}$\\
 Number of Point Clouds & 365297 \\ % 356297\\
 % Calibration Accuracy & \\
\bottomrule
\end{tabular}
\label{tab::dataset} 
\vspace{-10pt}
\end{table}

\label{sec::dataset}

%% file: content/experiment.tex
\section{Experiment Results}
We conduct three experiments using our \dataset~dataset to demonstrate its usefulness in off-road navigation and perception under degraded lighting conditions. 

\subsection{End-to-End Navigation Learning}
\begin{figure}
    \centering
    \includegraphics[width=1\linewidth]{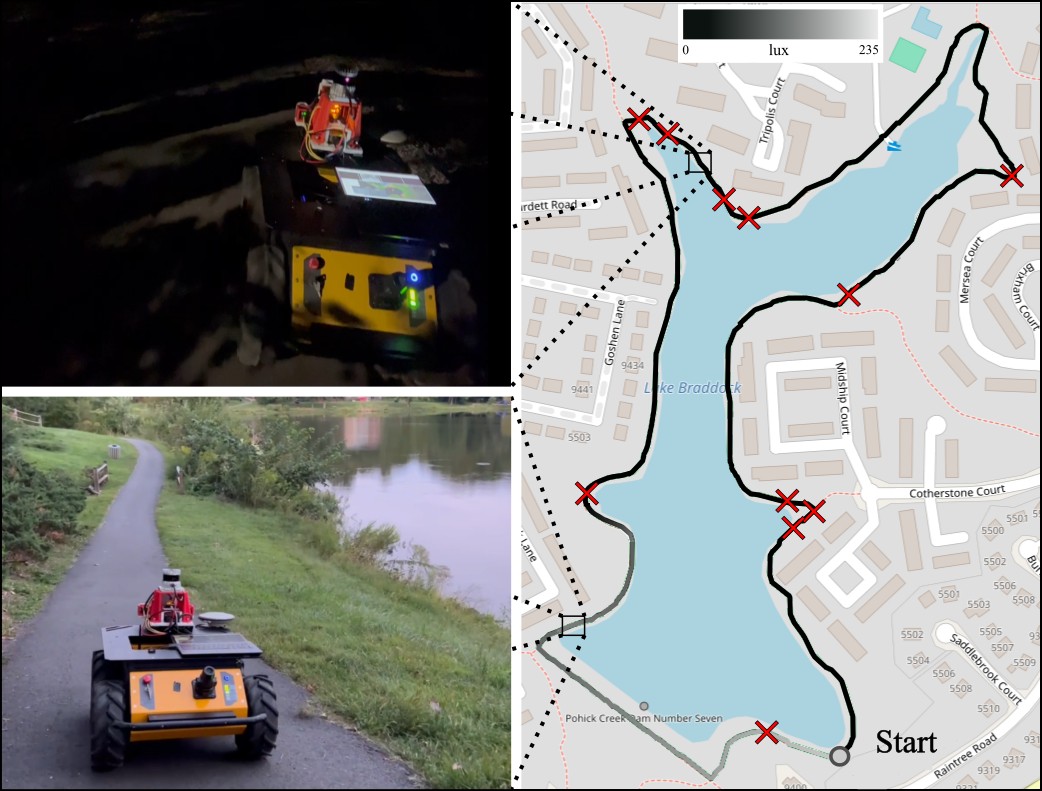}
    \caption{Autonomous Navigation around a 3.6 km Trail with a BC model and Thermal Input: Lighting conditions drops from 255 lx at the beginning (light gray on the path, lower right) to 0 lx (black, upper left). 11 interventions (red crosses) are necessary to correct the robot when going off-course. }
    \label{fig::bc_exp}
    \vspace{-5pt}
\end{figure}

To demonstrate the effectiveness of the dataset to enable end-to-end learning for autonomous navigation, we train an end-to-end behavior cloning (BC) model that outputs linear and angular velocities~\cite{datar2023toward, bojarski2016end} based on thermal camera input into a ResNet-18. Considering the difference in absolute temperature, we normalize each pixel value based on the max and min values of the current thermal image to get the relative temperature readings. We deploy this BC model on the Husky robot for a 3.6 km autonomous navigation task on a paved hiking trail, as illustrated in Fig.~\ref{fig::bc_exp}. The luminosity during the experiment ranges from 235 lx to 0 lx (indicated by the color of the path), with the robot completing most of the navigation in complete darkness (0 lx). The robot successfully completes the navigation, requiring only 11 human interventions when it goes off-course. Most interventions are because the pavement and the gravel on the side show similar temperature in the thermal input and therefore confuse the robot. More sophisticated techniques that leverage other sensor modalities, e.g., event camera, are necessary to enable more robust navigation. 

\subsection{Perception in Degraded Visual Environments}

\begin{table}
\centering
\caption{Quantitative Depth Prediction Comparison on Unseen Data.}
\begin{tabular}{lcccc}
\toprule
 Model & \#Params (M) $\downarrow$ & Abs Rel $\downarrow$& RMSE $\downarrow$& $\delta_1 \uparrow$ \\
 \midrule
  DepthAnythingV2 & 335.3 & 0.66 & 8.43 & 0.03\\
 U-Net + \dataset & 31 & 0.13 & 2.12 & 0.82\\
\bottomrule
\end{tabular}
\label{tab::depth_table} 
\vspace{-10pt}
\end{table}

To evaluate the efficacy of \dataset~in enabling scene perception in degraded visual environments, we conduct a comparative analysis of metric depth estimation. Specifically, we train a U-Net~\cite{ronneberger2015u}, 31M parameters, to learn a mapping between thermal infrared imagery and corresponding depth information derived from the LiDAR point clouds.
We compare the performance of this U-Net, trained on the M2P2 dataset, against DepthAnythingV2-Large~\cite{yang2024depth}, a monocular metric depth estimation model with approximately 335.3 million parameters. Quantitative results, detailed in Table.~\ref{tab::depth_table}, reveal a substantial performance superiority of U-Net despite its significantly lower parameter count. Notably, DepthAnythingV2-Large demonstrates limited generalization to the infrared domain.

\begin{figure}
    \centering
    \includegraphics[width=1\linewidth]{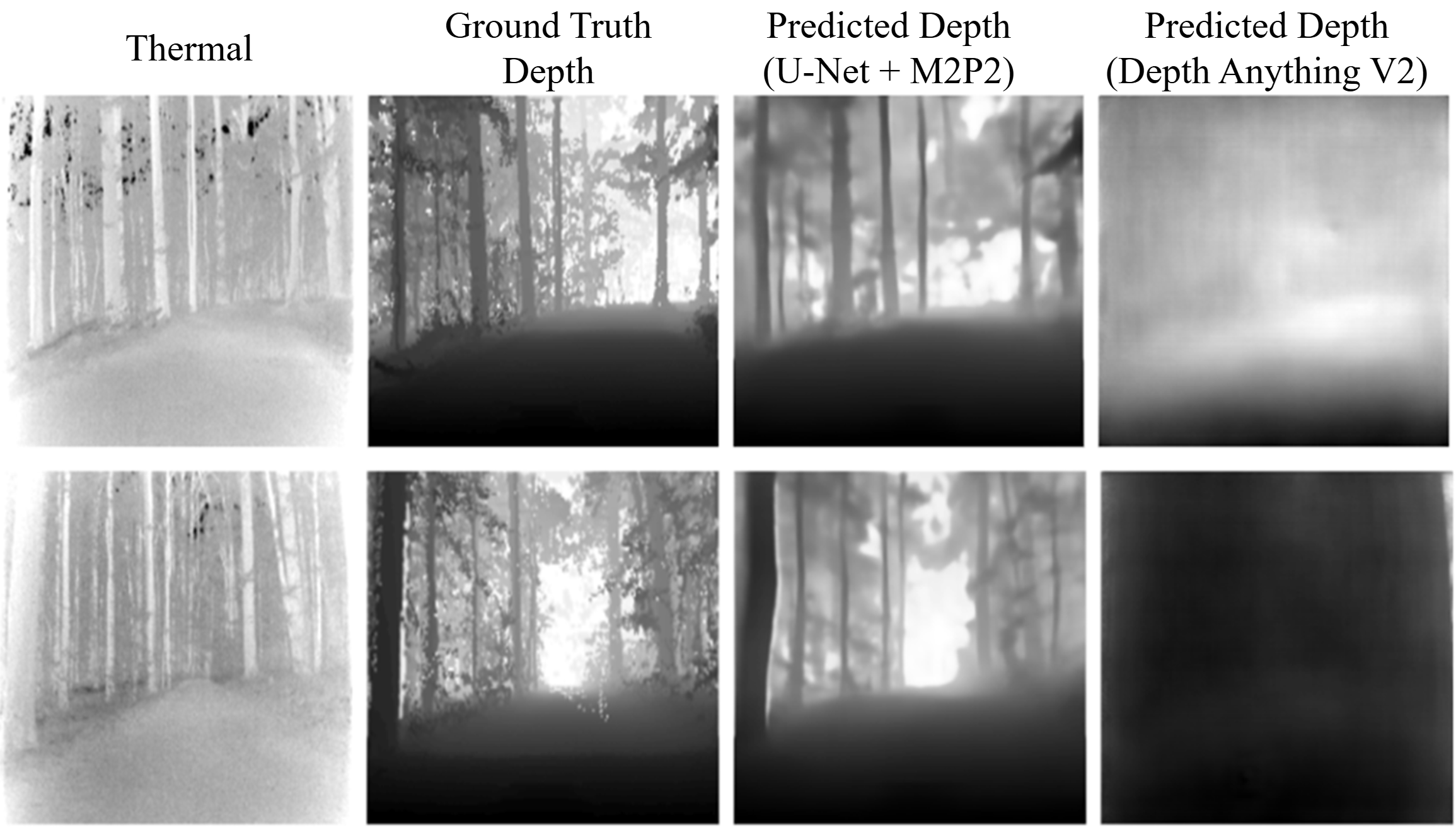}
    \caption{Qualitative Depth Prediction Comparison on Unseen Data.}
    \label{fig::depth}
    \vspace{-5pt}
\end{figure}

Qualitative evaluations, illustrated in Fig.~\ref{fig::depth}, further reinforce these findings. Qualitative inspection confirms that the U-Net trained on M2P2 generates depth maps of considerably higher fidelity compared to those produced by DepthAnythingV2-Large. This observation highlights the pivotal role of domain-specific datasets like M2P2 in enabling the development of robust perception models for degraded visual environments, where traditional RGB-based methods are inherently challenged. Our results suggest that such datasets are indispensable for bridging the gap between standard visual perception and the complexities introduced by atypical sensory inputs.

\subsection{Passive Visual Odometry with Thermal and Event Data}

A unique characteristic of M2P2 is the inclusion of calibrated, synchronized thermal and event camera data, enabling exploration of passive perception in extremely low-light conditions. While prior work has investigated visual-inertial odometry using RGB and event cameras~\cite{Pellerito_2024_IROS}, the fusion of thermal and event data for odometry remains relatively underexplored. This combination holds significant promise for applications where visible light is scarce or unavailable, such as nighttime off-road navigation or covert operations. The closest existing work is RAMP-VO~\cite{Pellerito_2024_IROS}, and M2P2 helps advance this area of research.

To demonstrate the potential of this multi-modal fusion, we adapt the RAMP-VO framework, originally designed for RGB and event data, to process thermal and event data from \dataset. We focus on a challenging 157.5 m segment of the Burke Lake trail to evaluate the robustness of the approach under varying light levels.

Crucially, we simulate reduced lighting conditions by systematically subsampling the event stream. This allows us to assess the performance of the thermal-event odometry system as the available information from the event camera decreases. We experiment with retaining 80\%, 50\%, and 25\% of the original events, representing progressively darker scenarios, in addition to using the full event data (100\%).

Table~\ref{tab::thermal_event_vo} presents the translational Absolute Trajectory Error (ATE) for each event subsampling level. As expected, the error generally increases as the event data becomes sparser.

\begin{table}
\centering
\caption{Translational ATE with Thermal-Event Fusion}
\begin{tabular}{cc}
\toprule
 Event Percentage & Translational ATE (m) $\downarrow$ \\
 \midrule
 100\% (Full Event Data) & 8.79 \\
 80\%  & 11.60 \\
 50\%  & 12.79 \\
 25\%  & 12.49 \\

\bottomrule
\end{tabular}
\label{tab::thermal_event_vo}
\vspace{-10pt}
\end{table}

\label{sec::experiments}

%% file: content/conclusion.tex
\section{Conclusions and Future Work}
\label{sec::conclusions}

This paper introduces M2P2, a novel multi-modal passive perception dataset specifically designed to address the challenges of off-road robot mobility in extreme low-light conditions. Unlike existing datasets, M2P2 uniquely combines thermal, event, and stereo RGB cameras, along with IMUs, GPS, and LiDAR for ground truth, providing a comprehensive representation of challenging off-road, low-light environments. We make the M2P2 dataset, along with our sensor suite design, publicly available to facilitate further research. We also present a robust multi-sensor calibration procedure, ensuring accurate data alignment across all modalities. Our initial experiments demonstrate that, even in complete darkness, off-road navigation, scene understanding, and vehicle state estimation are achievable using purely passive sensing.

While these initial experiments showcase the promise of individual modalities and limited fusion, the full realization of M2P2's potential requires deeper exploration of advanced sensor fusion techniques and their application to a wider range of mobility tasks. As the first step toward fully passive perception for off-road mobility in extreme low-light conditions, this work opens up a new avenue of future research. Some of the areas that could benefit from M2P2 include Visual Inertial Odometry~\cite{zhang2018tutorial,huai2022robocentric, bloesch2015robust}, SLAM~\cite{davison2007monoslam, mur2015orb, taketomi2017visual}, and off-road kinodynamics modeling~\cite{xiao2021learning, karnan2022vi, atreya2022high, datar2023learning, datar2024terrain, pokhrel2024cahsor}, all with the purely passive modalities available from our multi-modal sensor suite and dataset.